\documentclass[11pt,a4paper]{article}
\usepackage[hyperref]{acl2019}
\usepackage{times}
\usepackage{comment}
\usepackage{latexsym}
\usepackage{color}
\usepackage{tabularx}
\usepackage{graphicx}

\usepackage{amsmath}
\newcommand{\urlwofont}[1]{\urlstyle{same}\url{#1}}
\newcolumntype{Y}{>{\centering\arraybackslash}X} 

\usepackage{url}

\aclfinalcopy % Uncomment this line for the final submission
\def\aclpaperid{1911} %  Enter the acl Paper ID here

\newcommand{\fb}[1]{[FB:#1]}

\title{Identification of Tasks, Datasets, Evaluation Metrics, and Numeric Scores for Scientific Leaderboards Construction}

\author{Yufang Hou, Charles Jochim, Martin Gleize, Francesca Bonin and Debasis Ganguly \\
IBM Research -- Ireland \\
Dublin, Ireland \\
  \texttt{\{yhou|charlesj|mgleize|fbonin|debasga1\}@ie.ibm.com} \\}
\date{}

\begin{document}
\maketitle

\begin{abstract}

While the fast-paced inception of novel tasks and new datasets helps foster active research in a community towards interesting directions, keeping track of the abundance of research activity in different areas on different datasets is likely to become increasingly difficult. The community could greatly benefit from an automatic system able to summarize scientific results, e.g., in the form of a \textit{leaderboard}.
In this paper we  build two datasets and develop a framework (\emph{TDMS-IE}) aimed at automatically extracting \textit{task, dataset, metric} and \textit{score} from NLP papers, towards the automatic construction of leaderboards. Experiments show that our model outperforms several baselines by a large margin. 
Our model is a first step towards automatic leaderboard construction, e.g., in the NLP domain.
\end{abstract}

\section{Introduction}\label{sec:intro}
\begin{figure*}[t]
\begin{center}
\includegraphics[width=1.0\textwidth]{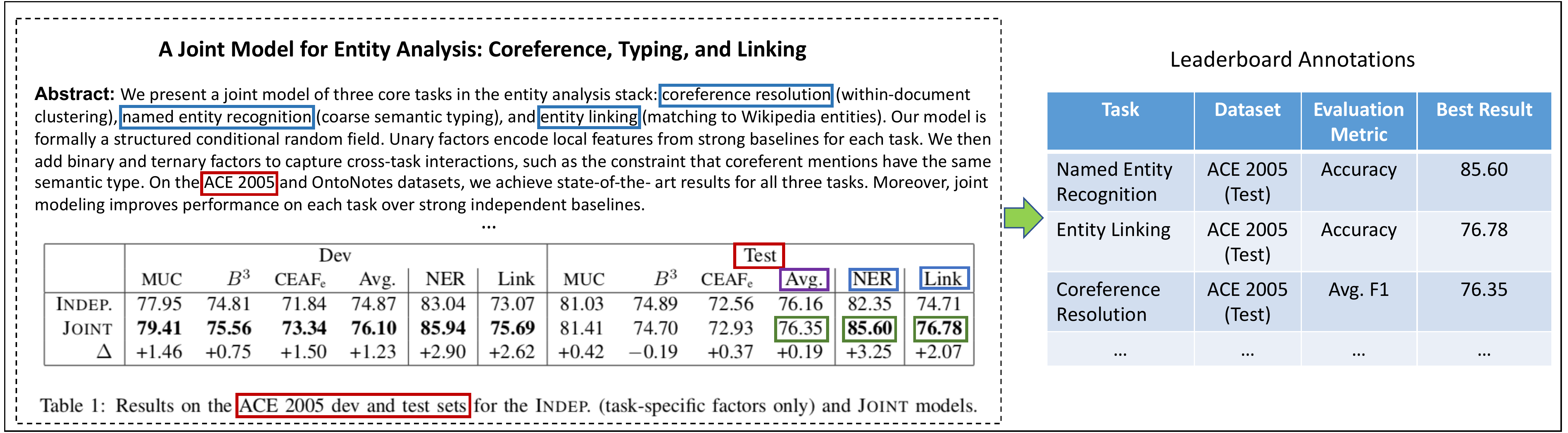}
\end{center}
\caption{An illustrative example of leaderboard construction from a sample article. The cue words related to the annotated tasks, datasets, evaluation metrics and the corresponding best scores are shown in blue, red, purple and green, respectively. 
Note that sometimes the cue words appearing in the article are different from the document-level annotations, e.g., Avg. -- Avg. F1, NER -- Named Entity Recognition.}
\label{fig:example}
\end{figure*}

Recent years have witnessed a significant increase in the number of laboratory-based evaluation benchmarks in many of scientific disciplines, e.g.,
in the year 2018 alone, 140,616 papers were submitted to the pre-print repository arXiv\footnote{\url{https://arxiv.org/}} and among them, 3,710 papers are under the \emph{Computer Science -- Computation and Language} category.
This massive increase in evaluation benchmarks (e.g., in the form of shared tasks) is particularly true for an empirical field such as NLP, which strongly encourages the research community to develop a set of publicly available benchmark tasks, datasets and tools so as to reinforce reproducible experiments.

Researchers have realized the importance of conducting meta-analysis of a number of comparable publications, i.e., the ones which use similar, if not identical, experimental settings, from shared tasks and proceedings, as shown by special issues dedicated to analysis of reproducibility in experiments \cite{Ferro:2018}, or by detailed comparative analysis of experimental results reported on the same dataset in published papers \cite{ArmstrongMWZ09}.

A useful output of this meta-analysis is often a summary of the results of a comparable set of experiments (in terms of the tasks they are applied on, the datasets on which they are tested and the metrics used for evaluation) in a tabular form, commonly referred to as a \emph{leaderboard}.
Such a meta-analysis summary in the form of a leaderboard is potentially useful to researchers for the purpose of (1) choosing the appropriate existing literature for fair comparisons against a newly proposed method; and (2) selecting strong baselines, which the new method should be compared against. 

Although recently there has been some effort to manually keep an account of progress on various research fields in the form of leaderboards,  either by individual researchers\footnote{\url{https://github.com/sebastianruder/NLP-progress}} or in a moderated crowd-sourced environment by organizations\footnote{\url{https://paperswithcode.com}}, it is likely to become increasingly difficult and time-consuming over the passage of time.

In this paper, we develop a model to automatically identify 
\emph{tasks}, \emph{datasets}, \emph{evaluation metrics}, and to extract the corresponding \emph{best numeric scores} from experimental scientific papers. An illustrative example is shown in Figure \ref{fig:example}: given the sample paper shown on the left, which carries out research work on three different tasks (i.e., \emph{coreference resolution}, \emph{named entity recognition}, and \emph{entity linking}), the system is supposed to extract the corresponding \emph{Task-Dataset-Metric-Score} tuples as shown on the right part in Figure \ref{fig:example}. 
It is noteworthy that we aim to identify a set of predefined \emph{Task-Dataset-Metric} (\emph{TDM}) triples from a taxonomy for a paper, and the corresponding cue words appearing in the paper could have a different surface form, e.g., \emph{Named Entity Recognition (taxonomy) -- Name Tagging (paper)}. 

Different from most previous work on information extraction from scientific literature which concentrates mainly on the abstract section or individual paragraphs \cite{semeval2017,semeval2018,Luan2018}, our task needs to analyze the entire paper. 
More importantly, our main goal is to \emph{tag} papers using TDM triples from a taxonomy and to use these triples to organize papers. 
%We cast this problem as a textual entailment task. 
We adopt an approach similar to that used for some natural language inference (NLI) tasks \cite{snli:emnlp2015,poliak:etal:18}.
Specifically, given a scientific paper in PDF format, our system first extracts the key contents from the abstract and experimental sections, as well as from the tables. 
%We treat these key contents as \emph{\textbf{text}} and the targeting 
Then, we identify a set of \emph{Task-Dataset-Metric} (\emph{TDM}) triples or \emph{Dataset-Metric} (\emph{DM}) pairs per paper.
Our approach predicts if the textual \emph{context} matches the \emph{TDM}/\emph{DM} label \emph{hypothesis}, forcing the model to learn the similarity patterns between the text and various \emph{TDM} triples. For instance, the model will capture the similarities between \emph{ROUGE-2} and ``Rg-2''. 
We further demonstrate that our framework is able to generalize to the new (unobserved)
TDM triples at test time in a zero-shot TDM triple identification setup.

To evaluate our approach, we create a dataset \emph{NLP-TDMS} which contains
around 800 leaderboard annotations for more than 300 papers. Experiments show that our model outperforms several baselines by a large margin for extracting \emph{TDM} triples. We further carry out experiments on a much larger dataset \emph{ARC-PDN} and demonstrate that our system can support the construction of various leaderboards from a large number of scientific papers in the NLP domain. 

To the best of our knowledge, our work is the first attempt towards the creation of NLP Leaderboards in an automatic fashion. We pre-process both datasets (papers in PDF format) using GROBID \cite{gorbid} and an in-house PDF table extractor.
The processed datasets and code are publicly available at: \url{https://github.com/IBM/science-result-extractor}.

\section{Related Work}
\label{sec:rel}

\begin{table*}[t]
\begin{center}
\begin{small}
\begin{tabular}{lccc}
\hline
&Macro P&Macro R&Macro F$_1$\\ \hline
\emph{Table caption}&79.2&87.0&82.6 \\
\emph{Numeric value +  IsBolded + Table caption}&71.1&77.7&74.0 \\
\emph{Numeric value +  Row label+ Table caption}&55.5&71.4&61.4 \\
\emph{Numeric value  + Column label + Table caption}&49.8&67.2&55.4 \\
\emph{Numeric value + IsBolded + Row label + Column label + Table caption }&36.6&60.9&43.0 \\
 \hline
\end{tabular}
\end{small}
\end{center}
\caption{\label{tab:tableExt}  Table extraction results of our table parser on 50 tables from 10 NLP papers in PDF format.}
\end{table*}

%\paragraph{Information extraction from scientific literature.}

A number of studies have recently explored methods for extracting information from scientific papers. 
Initial interest was shown in the analysis of citations \cite{athar-teufel:2012,athar-teufel:2012:DSSD,jurgens-etal-2018-measuring} and analysis of the topic trends in the scientific communities \cite{W12-3204}.   
\citet{gupta2011analyzing,GBOR16.870} propose unsupervised methods for the extraction of entities such as papers' focus and methodology; similarly, in \cite{Tsai:2013:}, an unsupervised bootstrapping method is used to identify and cluster the main concepts of a paper.
But only in 2017, \newcite{semeval2017} formalized a new task (SemEval 2017 Task 10) for the identification of three types of entities (called keyphrases, i.e., \textit{Tasks, Methods}, and \textit{Materials}) and two relation types (\textit{hyponym-of} and \textit{synonym-of}) in a corpus of 500 paragraphs from articles in the domains of \emph{Computer Science}, \emph{Material Sciences} and \emph{Physics}. \newcite{semeval2018} also presented the task of IE from scientific papers (SemEval 2018 Task 7) with a dataset of 350 annotated abstracts. 
 \citet{Ammar2017,Ammar2018,Luan2017,Augustein2017Acl} exploit these datasets to test neural models for IE on scientific literature.
\citet{Luan2018} extend those datasets by adding more relation types and cross-sentence relations using coreference links. The authors also develop a framework called Scientific Information Extractor for the extraction of six types of scientific entities (\textit{Task, Method, Metric, Material, Other-ScientificTerm} and \textit{Generic}) and seven relation types (\textit{Compare, Part-of, Conjunction, Evaluate-for, Feature-of, Used-for}, and \textit{Hyponym-of}). They reach 64.2 F$_1$ on entity recognition and 39.2 F$_1$ on relation extraction.
Differently from \cite{Luan2018}, (1) we concentrate on the identification of entities from a taxonomy that are necessary for the reconstruction of leaderboards (i.e., \textit{task, dataset, metric}); % and \textit{score}); 
(2) we analyse the entire paper, not only the abstract (the reason being that the \textit{score} information is rarely contained in the abstract).

%In order to tackle this last point, we propose a model for table parsing based on the library GROBID. Work on table extraction has been done in \cite{yildiz2005pdf2table,fang2012table} \fb{find more and more recent}; however, to the best of our knowledge, none of the previous works on table extraction from scientific papers is tailored to the extraction of the result. 

% \paragraph{\textbf{Textual entailment.}}
%Differently from previous approaches to extract information from scientific literature, we use a Textual Entailment framework for our task. Recognising Textual entailment (RTE) is the task of determining, given a premise and a hypothesis, whether the premise entails the hypothesis.
Our method for TDMS identification resembles some approaches used for textual entailment \cite{dagan:etal:06} or natural language inference (NLI) \cite{snli:emnlp2015}.
%RTE has recently captured the attention of many thanks to the generation of new datasets (SNLI, Multi-SNLI and Swag) \cite{snli:emnlp2015, zellers2018swag}. 
We follow the example of \citet{white:etal:17} and \citet{poliak:etal:18} who reframe different NLP tasks, including extraction tasks, as NLI problems.
\citet{eichler:etal:17} and \citet{obamuyide:vlachos:18} have both used NLI approaches for relation extraction.
Our work differs in the information extracted and consequently in what \emph{context} and \emph{hypothesis} information we model.
Currently, one of the best performing NLI models (e.g., on the SNLI dataset) for three way classification is \cite{DBLP:journals/corr/abs-1901-11504}. The authors apply deep neural networks and make use of BERT \cite{BERT}, a novel language representation model. They reach an accuracy of 91.1\%. \citet{DBLP:journals/corr/abs-1805-11360} exploit densely-connected co-attentive recurrent neural network, and reach 90\% accuracy. 
In our scenario, we generate pseudo 
premises and hypotheses, then apply the standard transformer encoder \cite{ashish17,BERT} to train two NLI models.

\begin{comment}
\paragraph{\textbf{Automatic leaderboard construction}}
\fb{@all decide in the end if keep this part of no}
The final aim of our research line is the automatic construction of leaderboards in the NLP domain. Some efforts have been done in keeping an account of research activities either manually by individual researchers\footnote{\url{https://github.com/sebastianruder/NLP-progress}} or in a moderate crowdsourced fashion by organizations\footnote{\url{https://paperswithcode.com/}}. However, to the best of our knowledge, our work is the first attempt towards the creation of NLP Leaderboards in an automatic fashion. \citet{Blum2015} have tackled the issue from a different perspective: the authors propose an evaluation metric for machine learning competitions that faithfully represents the quality of the best submission. Our interest, instead, is extracting information reported in papers in order to build leaderboard based on such results.
\end{comment}
 
 %\bigskip
 \begin{comment}
To summarise: differently from all the previous approaches, we: \textit{1)} extract information from the entire article, not only abstracts; \textit{2)} we provide a method for extracting the \textit{score} of a paper, in addition to \textit{task, dataset} and \textit{metric}; \textit{3)} we propose a framework inspired on textual entailment;
 \textit{4)} we exploit the information of the tables, that include essential information on the papers' score. 
\end{comment}

%%% Local Variables:
%%% mode: latex
%%% TeX-master: "nlp-leaderboard"
%%% End:

\section{Dataset Construction}
\label{sec:datasets}

We create two datasets for testing our approach for \emph{task, dataset, metric, and score} (TDMS) identification.
Both datasets are taken from a collection of NLP papers in PDF format and both require similar pre-processing.
First, we parse the PDFs using GROBID \cite{gorbid} to extract the title, abstract, and for each section, the section title and its corresponding content. 
Then we apply an improved table parser we developed, built on GROBID's output, to extract all tables containing numeric cells from the paper.
%\fb{TODO potentially present PDF parser performance (probably)}
Each extracted table contains the table caption and a list of numeric cells. For each numeric cell, we detect whether it has a bold typeface, and associate it to its corresponding row and column headers. For instance, for the sample paper shown in Figure \ref{fig:example}, after processing the table shown, we extract the bolded number ``85.60'' and find its corresponding column headers ``\{Test, NER\}''.

We evaluated our table parser on a set of 10 papers from different venues (e.g., \emph{EMNLP, Computational Linguistics journal}).
In total, these papers contain 50 tables with 1,063 numeric content cells. Table \ref{tab:tableExt} shows the results for extracting different table elements.  Our table parser achieves a macro F$_1$ score of 82.6 for identifying table captions, and 74.0 macro F$_1$  for extracting tuples of $<$\emph{Numeric value, Bolded Info, Table caption}$>$. In general, it obtains higher recall than precision in all evaluation dimensions.

In the remainder of this section we describe our two datasets in
detail.
%\footnote{The datasets, including the extracted tables, will be available on publication.}

\subsection{NLP-TDMS}
\label{sec:ds1}
The content of the NLP-progress Github repository\footnote{\url{https://github.com/sebastianruder/NLP-progress}} provides us with expert annotations of various leaderboards for a few hundred papers in the NLP domain. The repository is organized following a ``language-domain/task-dataset-leaderboard'' structure. After crawling this information together with the corresponding papers (in PDF format), we clean the dataset manually. This includes: (1) normalizing task name, dataset name, and evaluation metrics across  leaderboards created by different experts, e.g., using ``F1'' to represent ``F-score'' and ``Fscore''; (2) for each leaderboard table, only keeping the best result from the same paper\footnote{In this paper, we focus on tagging papers with different leaderboards (i.e., \emph{TDM} triples). For each leaderboard table, an ideal situation would be to extract all results reported in the same paper and associate them to different \emph{methods}, we leave this for future work.}; (3) splitting a leaderboard table into several leaderboard tables 
if its column headers represent datasets instead of evaluation metrics. 

The resulting dataset \emph{NLP-TDMS (Full)} contains 332 papers with 848 leaderboard annotations. Each leaderboard annotation is a tuple containing \emph{task, dataset, metric}, and \emph{score} (as shown in Figure \ref{fig:example}). In total, we have 168 distinct leaderboards (i.e., $<$\emph{Task, Dataset, Metric}$>$ triples) and only around half of them (77) are associated with at least five papers. We treat these manually curated  \emph{TDM} triples as an NLP knowledge taxonomy and we aim to explore how well we can associate a paper to the corresponding \emph{TDM} triples.

\begin{table}[t]
\begin{center}
%\begin{small}
\begin{tabular}{@{}lcc@{}}
%&\multicolumn{2}{c}{\emph{NLP-TDES}}\\ \hline
%\hline
%&NLP-TDMS&NLP-TDMS\\ 
&Full&Exp\\ \hline
Papers   &332&332\\ %\hline 
Extracted tables   &1269&1269\\ %\hline 
``Unknown'' annotations &-&90\\ \hline 
Leaderboard annotations &  848&606\\ 
\enspace\enspace Distinct leaderboards& 168&77\\
\enspace\enspace Distinct tasks & 35&18\\
\enspace\enspace Distinct datasets &99&44\\
\enspace\enspace Distinct metrics & 72&30\\
 \hline
\end{tabular}
%\end{small}
\end{center}
\caption{\label{tab:ds1}  Statistics of leaderboard annotations in \emph{NLP-TDMS (Full)} and \emph{NLP-TDMS (Exp)}.}
\end{table}

We further create \emph{NLP-TDMS (Exp)} by removing those leaderboards that are associated with fewer than five papers. If all leaderboard annotations of a paper belong to 
these removed leaderboards, we tag this paper as ``Unknown''. Table \ref{tab:ds1} compares 
statistics of \emph{NLP-TDMS (Full)} and \emph{NLP-TDMS (Exp)}.
All experiments in this paper (except experiments in the zero-shot setup in Section \ref{sec:zeroshot_results}) are on \emph{NLP-TDMS (Exp)} and going forward we will refer to that only as \emph{NLP-TDMS}.
%The 77 leaderboards of this dataset constitute the set of TDM labels we aim to predict. 

\subsection{ARC-PDN}
\label{sec:ds2}
To test our model in a more realistic scenario, we create a second dataset \emph{ARC-PDN}.\footnote{PDN comes from the anthology's directory prefixes for ACL, EMNLP, and NAACL, respectively.}
We select papers (in PDF format) published in ACL, EMNLP, and NAACL between 2010 to 2015 from the most recent version of the ACL Anthology Reference Corpus (ARC) \cite{L08-1005}.
Table \ref{tab:ds2} shows statistics about papers and extracted tables in this dataset after the PDF parsing described above.

\begin{table}[t]
\begin{center}
\begin{tabular}{lcc}
%\hline
&\#Papers&\#Extracted tables \\ \hline
 ACL&1958&4537\\ \hline 
 EMNLP&1167&3488\\ \hline 
 NAACL&730&1559\\ \hline 
 Total&3855&9584\\ \hline 
\end{tabular}
\end{center}
\caption{\label{tab:ds2}  Statistics of papers and extracted tables in \emph{ARC-PDN}.}
\end{table}

%%% Local Variables:
%%% mode: latex
%%% TeX-master: "nlp-leaderboard"
%%% End:

\section{Method for TDMS Identification}
\label{sec:method}
%Experiment scientific papers in the CS domain often focus on proposing new method to
%address 

\subsection{Problem Definition}
\label{sec:problemDef}
We represent each leaderboard as a $<$\emph{Task, Dataset, Metric}$>$ triple (\emph{TDM} triple). Given an experimental scientific paper $D$, we
want to identify relevant \emph{TDM} triples from a taxonomy and extract %along with   
the best numeric score for each predicted \emph{TDM} triple.
 
\begin{figure*}[t]
\begin{center}
\includegraphics[width=0.99\textwidth]{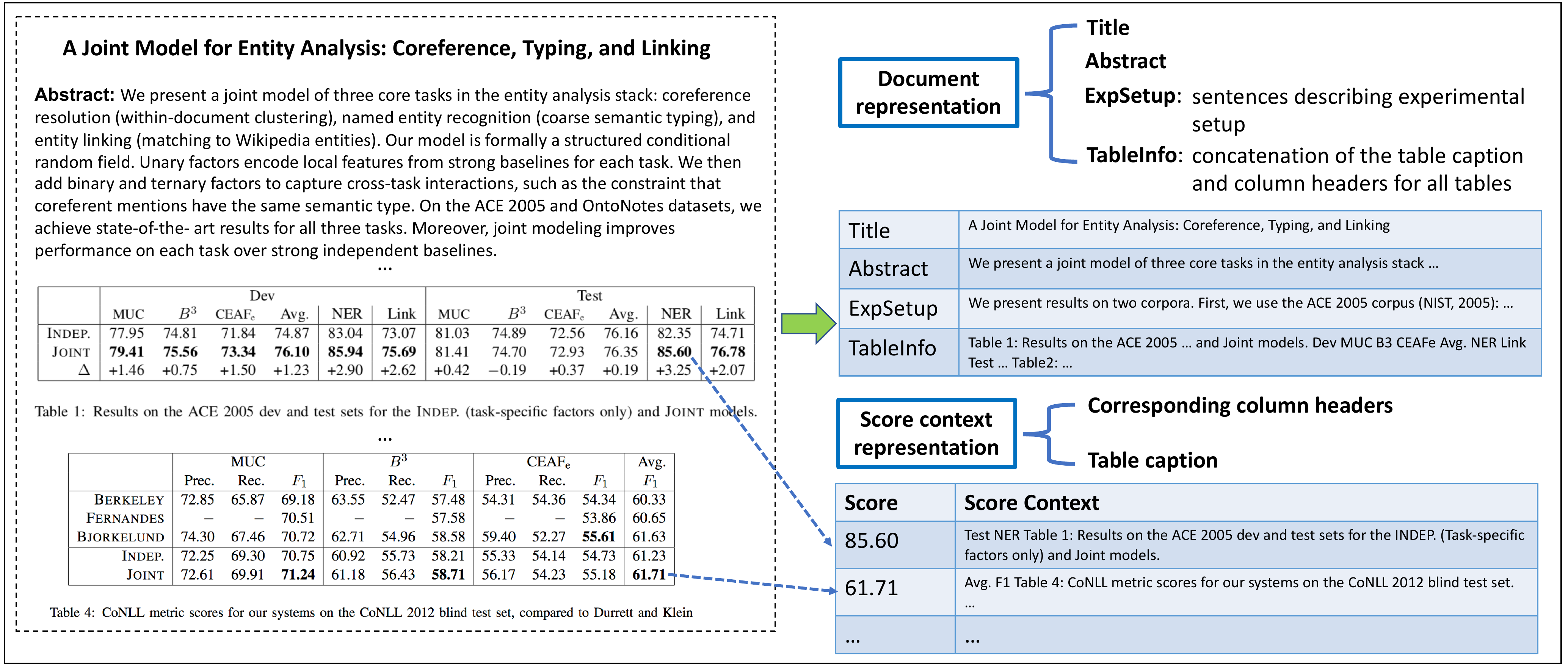}
\end{center}
\caption{Examples of document representation (\emph{DocTAET}) and score context (\emph{SC}) representation.}
\label{fig:docrepresentation}
\end{figure*}

However, scientific papers are often long documents and only some parts of the document are useful to predict \emph{TDM} triples and the associated scores. 
%Here we define document representation \emph{DocTAET} and table score context \emph{SC} for scientific papers as follows:
Hence, we define a document representation, called  \emph{DocTAET} and a table \textit{score} representation, called \emph{SC} (score context), as follows:

\paragraph{DocTAET.} For each scientific paper, its \emph{DocTAET} representation contains the following four parts: \emph{Title, Abstract, ExpSetup}, and \emph{TableInfo}.
\emph{Title} and \emph{Abstract} often help in predicting \emph{Task}. \emph{ExpSetup} contains all sentences which are likely to describe the experimental setup, which can help to predict \emph{Dataset} and \emph{Metric}. We use a few 
heuristics to extract such sentences.\footnote{A sentence is included in \emph{ExpSetup} if it: (1) contains any of the following cue words/phrases:
\{\emph{experiment on, experiment in, evaluation(s), evaluate, evaluated, dataset(s), corpus, corpora}\}; and (2) belongs to a section whose title contains any of the following words: \{\emph{experiment(s)}, \emph{evaluation}, \emph{dataset(s)}\}.}
Finally, table captions and column headers are important in predicting \emph{Dataset} and \emph{Metric}. We collect them in the \emph{TableInfo} part. 
Figure \ref{fig:docrepresentation} (upper right) illustrates the \emph{DocTAET} extraction for a given paper.
%concatenates the table caption 
%with its column headers for all tables.   

\paragraph{SC.} For each table in a scientific paper, we focus on boldfaced numeric scores because they are more likely to be the best scores for the corresponding \emph{TDM} triples.\footnote{We randomly choose 10 papers from \emph{NLP-TDMS (Full)} and compare their \emph{TDMS} tuple annotations with the results reported in the original tables.  We found that 78\% (18/23) of the annotated tuples contain boldfaced numeric scores.} For a specific boldfaced numeric score in a table, its context (\emph{SC}) contains its corresponding column headers and the table caption. Figure \ref{fig:docrepresentation} (lower right) shows the extracted \emph{SC} for the scores \emph{85.60} and \emph{61.71}.
%two scores, i.e., \emph{85.60} and \emph{61.71} respectively.

\subsection{TDMS-IE System}
\label{sec:tdms-ie}
\begin{figure*}[t]
\begin{center}
\includegraphics[width=0.99\textwidth]{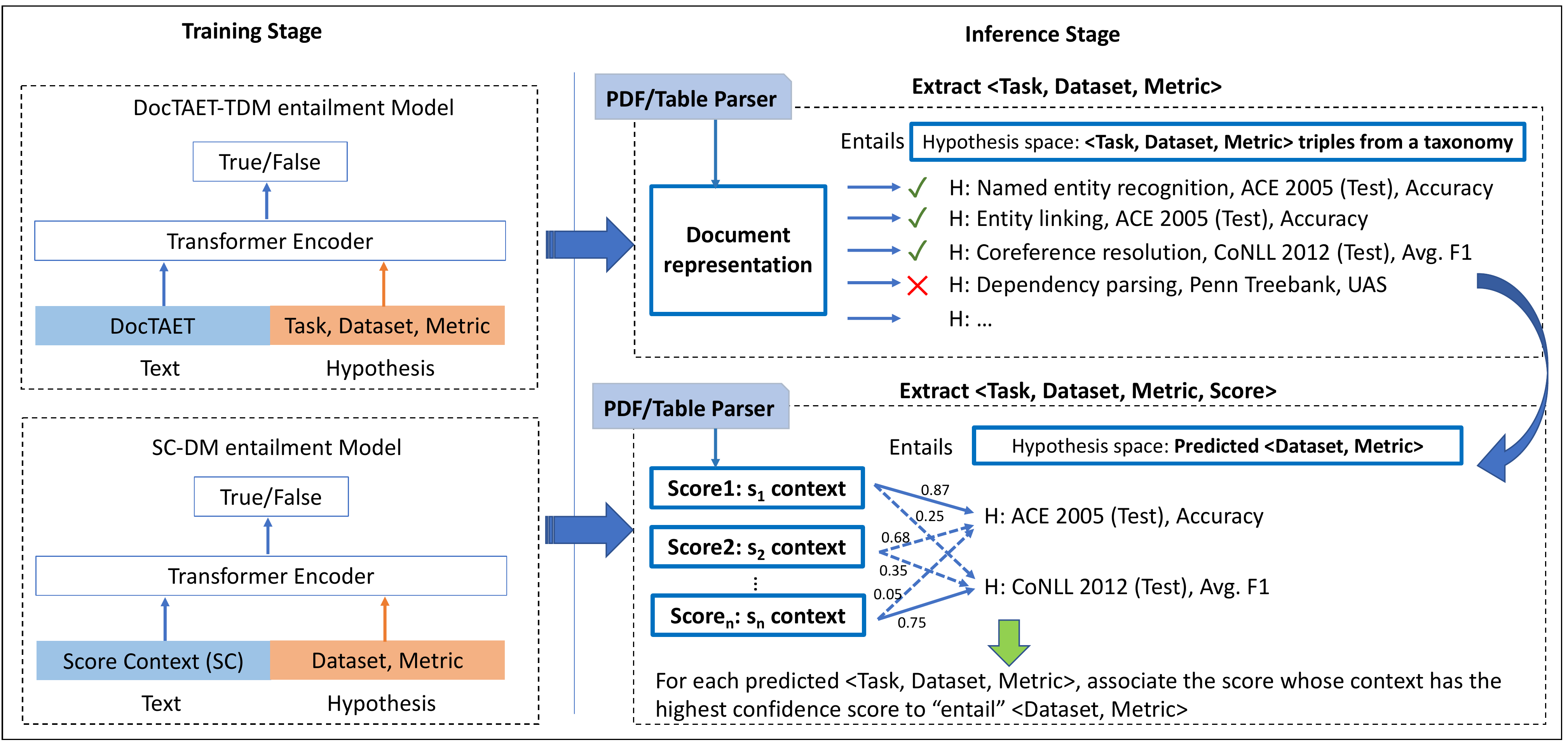}
\end{center}
\caption{System architecture for \emph{TDMS-IE}.}
\label{fig:system}
\end{figure*}

We develop a system called \emph{TDMS-IE} to associate \emph{TDM} triples to a given experimental scientific paper. Our system also extracts the best numeric \textit{score} for each predicted \emph{TDM} triple. Figure \ref{fig:system} shows the system architecture for \emph{TDMS-IE}. 

\subsubsection{TDMS-IE Classification Models}
\label{sec:tdms-ie-train}
To predict correct \emph{TDM} triples and associate the appropriate scores, we adopt a natural language inference approach (NLI) \cite{poliak:etal:18} and learn a binary classifier for pairs of document \emph{contexts} and \emph{TDM} label \emph{hypotheses}.
%We formulate the task as a textual entailment problem. 
Specifically, we split the problem into two tasks:  (1) given a document representation \emph{DocTAET}, we would like to predict whether a specific \emph{TDM} triple can be inferred (e.g., give a document we infer $<$\emph{Summarization, Gigaword, ROUGE-2}$>$); 
%\footnote{UAS means unlabeled attachment score and is an evaluation metric for dependency parsing.}\}). 
(2) we predict whether a $<$\emph{Dataset, Metric}$>$ tuple (\emph{DM}) can be inferred given a score context \emph{SC}.\footnote{We look for the relation \emph{SC}-\emph{DM}, rather then \emph{SC}-\emph{TDM}, because rarely the task is mentioned in \emph{SC}.}
This setup has two advantages: first, it naturally captures the inter-relations between different labels by encoding the three types of labels (i.e., task, dataset, metric) into the same hypothesis. 
%For instance,  if a document representation has strong signals to predict \emph{Penn Treebank} and \emph{UAS}, it will predict the task label \emph{Dependency parsing} as well.
Second, similar to approaches for NLI, it forces the model to focus on learning the similarity patterns between \emph{DocTAET} and various \emph{TDM} triples. For instance, the model will capture the similarities between \emph{ROUGE-2} and ``Rg-2''.

Recently, a multi-head self-attention encoder \cite{ashish17} has been shown to perform well in various NLP tasks, including NLI \cite{BERT}. 
We apply the standard transformer encoder \cite{BERT} to train our models, one for \emph{TDM} triple prediction, and one for score extraction. In the following we describe how we generate training instances for these two models.

\paragraph{DocTAET-TDM model.} Illustrated in Figure \ref{fig:system} (upper left), this model predicts whether a \emph{TDM} triple can be inferred from a \emph{DocTAET}. For a set of $n$ \emph{TDM} triples (\{$t_1, t_2, ..., t_n$\}) from a taxonomy, if a paper $d_i$ (\emph{DocTAET}) is annotated with $t_1$ and $t_2$, we then generate two positive training instances ($d_i \Rightarrow t_1$ and $d_i \Rightarrow t_2$) and $n-2$ negative training instances ($d_i \not \Rightarrow t_j$, $2<j \leq n$).
%Once trained, this model can then predict which \emph{TDM} triples are entailed by a new paper.

\paragraph{SC-DM model.} Illustrated in Figure \ref{fig:system} (lower left), this model predicts whether a score context \emph{SC} indicates a \emph{DM} pair. 
To form training instances, we start with the list of \emph{DM} pairs (\{$p_1, p_2, ..., p_m$\}) from a taxonomy and a paper $d_i$,  which is annotated with a \emph{TDM} triple $t$ (containing  $p_1$) and a numeric score $s$. We first try to extract the score contexts (\emph{SC}) for all bolded numeric scores. If $d_i$'s annotated score $s$ is equal to one of the bolded scores $s_k$ (typically there should not be more than one), we generate a positive training instance ($SC_{s_{k=1}} \Rightarrow p_1$).  Negative instances can be generated for this context by choosing other $DM$s not associated with the context, i.e.,  $m-1$ negative training instances ($SC_{s_{k=1}} \not \Rightarrow p_j$, $1<j \leq m$).
For example, an $SC$ with ``ROUGE for anonymized CNN/Daily Mail'' might form a positive instance with $DM$ $<$\emph{CNN / Daily Mail, ROUGE-L}$>$, and then a negative instance with $DM$ $<$\emph{Penn Treebank, LAS}$>$.
Additional negative training instances come from bolded scores $s_k$ which do not match $s$ (e.g., $SC_{s_k} \not \Rightarrow p_j$, $1< k$, $1 \leq j \leq m$). 

\subsubsection{Inference}
During the inference stage (see Figure \ref{fig:system} (right)), for a given scientific paper
in PDF format, our system first uses the PDF parser and table extractor (described in Section \ref{sec:datasets}) to generate the document representation \emph{DocTAET}. We also extract all boldfaced scores and their contexts from each table. Next, we apply the \emph{DocTAET-TDM} model to predict \emph{TDM} triples among all \emph{TDM}  triple candidates for the paper\footnote{The \emph{TDM} triple candidates could be the valid \emph{TDM}  triples from the training set, or a set of \emph{TDM} triples from a taxonomy.}. Then, to extract scores for the predicted \emph{TDM} triples, we apply the \emph{SC-DM} model to every extracted score context (\emph{SC}) and predicted \emph{DM} pair (taken from the predicted \emph{TDM} triples). This step tells us how likely it is that a score context suggests a \emph{DM} pair. Finally, for each predicted \emph{TDM} triple, we select the score whose context has the highest confidence in predicting a link to the constituent \emph{DM} pair.

%%% Local Variables:
%%% mode: latex
%%% TeX-master: "nlp-leaderboard"
%%% End:

\section{Experimental Setup}\label{sec:exp_setup}
\subsection{Training/Test Datasets}
We split \emph{NLP-TDMS} (described in Section \ref{sec:datasets}) into training and test sets. 
The partitioning ensures that every \emph{TDM} triple annotated in \emph{NLP-TDMS} appears both in the training and test set, so that a classifier will not have to predict unseen labels (or infer unseen hypotheses).
%Papers can appear in multiple leaderboards, so particular care was taken to be sure no papers were in both training and test sets.
Table \ref{tab:exp1stat} shows statistics on these two splits.
The 77 leaderboards in this dataset constitute the set of $n$ \emph{TDM} triples we aim to predict (see Section \ref{sec:tdms-ie}). 

For evaluation, we report macro- and micro-averaged precision, recall, and F$_1$ score for extracting \emph{TDM} triples and \emph{TDMS} tuples over papers in the test set.
\begin{comment}
\fb{cj: not sure the following is needed}
Since this output could ultimately be used to generate a leaderboard, i.e., a ranked list of papers ordered by performance, it seems natural to measure how close our generated leaderboards are to manually curated ones using some rank correlation (e.g., Spearman's $\rho$ or Kendall's $\tau$).
However, rank correlations are typically used when the order of the ranking is defined by the system, e.g., some relevance, similarity, or confidence score. 
In this work, the score is extracted from the paper and not any reflection of our system's confidence.
With a rank correlation it would be possible to have two systems that both make one error in labeling, but that get different scores because of the arbitrary position in which the mistake was made.
\end{comment}
% \footnote{For example in a list with five entries if system $A$ mislabels the first paper in the gold ranking and system $B$ mislabels the last, $B$ will have a higher rank score even though both systems only mislabeled one paper.}
%To evaluate performance on this task it seems more appropriate to use precision, recall, and F$_1$ to measure exactly when we are extracting the TDM(S) labels correctly.
%That being said, while we do not compare systems using Kendall's $\tau$, it is still interesting to apply it to see how close we are to the leaderboard.

\subsection{Implementation Details}\label{sec:nn_detail}
Both of our models (\emph{DocTAET-TDM} and \emph{SC-DM}) have 12 transformer blocks, 768 hidden units, and 12 self-attention heads.
For \emph{DocTAET-TDM},
we first  initialize it using BERT$_{BASE}$, then fine-tune the model for 3 epochs with the learning rate of $5e-5$.
During training and testing,  the maximum text length is set to 512 tokens.
Note that the document representation \emph{DocTAET} can contain more than 1000 tokens for some scientific papers, often due to very long content in \emph{ExpSetup} and \emph{TableInfo}. 
Therefore, in these cases, we use only the first 150 tokens from \emph{ExpSetup} and \emph{TableInfo} respectively.

\begin{table}[t]
\begin{center}
%\begin{small}
\begin{tabular}{@{}lcc@{}}
%&\multicolumn{2}{c}{\emph{NLP-TDES}}\\ \hline
%\hline
%&NLP-TDMS&NLP-TDMS\\
&training&test\\ \hline
Papers   &170&162\\ %\hline
Extracted tables   &679&590\\ %\hline
``Unknown'' annotations &46&44\\ \hline
Leaderboard annotations &  325&281\\
\enspace\enspace Distinct leaderboards& 77&77\\
 \hline
\end{tabular}
%\end{small}
\end{center}
\caption{\label{tab:exp1stat}  Statistics of training/test sets in \emph{NLP-TDMS}.}
\end{table}

We initialize the \emph{SC-DM} model using the trained \emph{DocTAET-TDM} model.
We suspect that 
\emph{DocTAET-TDM} already captures some of the relationship between score contexts and \emph{DM} pairs. After initialization, we continue
fine-tuning the model for 3 epochs with the learning rate of $5e-5$.
For \emph{SC-DM}, we set a maximum token length of 128 for both training and testing.

\subsection{Baselines}\label{sec:baselines}
In this section, we introduce three baselines against which we can evaluate our method. %one heuristics-based (StringMatch) and one using machine learning.

\paragraph{StringMatch (SM).} 
Given a paper, for
each \emph{TDM} triple, we first check whether the content of the title, abstract, or introduction contains the name of the task. Then we inspect the contexts of all extracted boldfaced scores to check whether: (1) the name of the dataset is mentioned in the table caption and one of the associated column headers matches the metric name; or (2) the metric name is mentioned in the table caption and one of the associated column headers matches the dataset name. If more than one numeric score is identified during the previous step, we choose the highest or lowest value according to the property of the metric (e.g., \emph{accuracy} should be high, while \emph{perplexity} should be low).

Finally, if all of the above conditions are satisfied for a given paper, we predict the \emph{TDM} triple along with the chosen score. 
Otherwise, we tag the paper as ``Unknown''.

%%% Local Variables:
%%% mode: latex
%%% TeX-master: "nlp-leaderboard"
%%% End:
\begin{table*}[!ht]
\begin{center}
\begin{small}
%\begin{tabular}{lcccccc}
\begin{tabularx}{\textwidth}{lYYYYYY}
\hline
&Macro P&Macro R&Macro F$_1$&Micro P&Micro R&Micro F$_1$\\ \hline
\multicolumn{7}{c}{(a) Task + Dataset + Metric Extraction}\\ \hline
\emph{SM}&31.8&30.6&31.0&36.0&19.6&25.4 \\
\emph{MLC}&42.0&23.1&27.8&42.0&20.9&27.9 \\
\emph{EL}& 18.1 & 31.8 & 20.5 & 24.3 & 36.3 & 29.1 \\
\emph{TDMS-IE}&\textbf{62.5}&\textbf{75.2}&\textbf{65.3}&\textbf{60.8}&\textbf{76.8}& \textbf{67.8}\\
 \hline
\multicolumn{7}{c}{(b) Task + Dataset + Metric Extraction (excluding papers with ``Unknown'' annotation)}\\ \hline
\emph{SM}&8.1&6.4&6.9&16.8&7.8& 10.6\\
\emph{MLC}&\textbf{56.8}&30.9&37.3&56.8&23.8&33.6 \\
\emph{EL}& 24.9 & 43.6 & 28.1 & 29.4 & 42.0 & 34.6 \\
\emph{TDMS-IE}&54.1&\textbf{65.9}&\textbf{56.6}&\textbf{60.2}&\textbf{73.1}&\textbf{66.0} \\
 \hline
\multicolumn{7}{c}{(c) Task + Dataset + Metric + Score Extraction (excluding papers with ``Unknown'' annotation)}\\ \hline
\emph{SM}&1.3&1.0&1.1&3.8&1.8&2.4 \\
\emph{MLC}&6.8&6.1&6.2&6.8&2.9&4.0 \\
\emph{TDMS-IE}&9.3&11.8&9.9&10.8&13.1&11.8 \\
 \hline
\end{tabularx}
\end{small}
\end{center}
\caption{\label{tab:exp1}  Leaderboard extraction results of \emph{TDMS-IE} and several baselines on the \emph{NLP-TDMS} test dataset.}
\end{table*}

\paragraph{Multi-label classification (MLC).}
For a machine learning baseline, we treat this task as a multi-class, multi-label classification problem where we would like to predict the \emph{TDM} label for a given paper (as opposed to predicting whether we can infer a given \emph{TDM} label based on the paper).
The class labels are \emph{TDM} triples and each paper can have multiple \emph{TDM} labels as they may report results from different tasks, datasets, and with different metrics.
For this classification we ignore instances with the `Unknown' label in training because this does not form a coherent class (and would otherwise dominate the other classes).
Then, for each paper, we extract bag-of-word features with tf-idf weights from the DocTAET representation described in Section \ref{sec:method}.
We train a multinomial logistic regression classifier implemented in scikit-learn \cite{scikit-learn}
%\footnote{\url{https://scikit-learn.org}} 
using SAGA optimization \cite{defazio:etal:14}.
In this multi-label setting, the classifier can return an empty set of labels.
When this is the case we take the most likely \emph{TDM} label as the prediction.

After predicting \emph{TDM} labels we need a separate baseline classifier to compare to the SC-DM model.
Similar to the SC-DM model, the MLC should predict the best score based on the \emph{SC}.
For training this classifier we form instances from triples of paper, score, and \emph{SC} (as described in Section \ref{sec:method}), with a binary label for whether or not this score is the actual leaderboard score from the paper.
This version of the training set for classification has 1647 instances, but is quite skewed with only 67 \emph{true} labels.
This skew is not as problematic because for this baseline we are not classifying whether or not the SC matches the leaderboard score, but instead we simply pick the most likely SC for a given paper.\footnote{Papers in the test set have an average of 47.3 scores to choose between.}
% 91 papers in test, 4303 lines
The scores chosen (in this case one per paper) are combined with the \emph{TDM} predictions above to form the final \emph{TDMS} predictions reported in Section \ref{sec:nlptdes_results}. 

\paragraph{EntityLinking (EL) for \emph{TDM} triples prediction.} We apply the state-of-the-art IE system on scientific literature \cite{Luan2018} 
to extract \emph{task}, \emph{material} and \emph{metric} mentions from \emph{DocTAET}. We then generate possible \emph{TDM} triples by combining these three types of mentions (note that many combinations could be invalid \emph{TDM} triples). Finally we link these candidates to the valid \emph{TDM} triples in a taxonomy\footnote{In this experiment, the taxonomy consists of 77 \emph{TDM} triples reported in Table \ref{tab:exp1stat}.} based on Jaccard similarity.  %and tag papers using the standard \emph{TDM} triples from the taxonomy. 
Specifically, we predict a \emph{TDM} triple for a paper if the similarity score between the triple and a candidate is greater than $\alpha$ ($\alpha$ is estimated in the training set). If none of \emph{TDM} triples was identified, we tag the paper as ``Unknown''.

%% also tested classifying TDM separately, and predicting labels independently and jointly
%In fact, we separate the task into three classification problems, one for task, dataset, and metrics, respectively.
%Each of these can still be considered a multi-class, multi-label problem, e.g., a paper may report multiple metrics, F$_1$ and accuracy.
%We treat the labels independently and concatenate the predicted labels from the respective classifiers to make the final prediction.

%% weaknesses of this baseline, can't really predict 'unknown'.  

%---

%%% Local Variables: 
%%% mode: latex
%%% TeX-master: "nlp-leaderboard"
%%% End: 
\section{Experimental Results}\label{sec:results}

\begin{table*}[t]
\begin{center}
\begin{small}
\begin{tabular}{lcccccc}
\hline
Document Representation&Macro P&Macro R&Macro F$_1$&Micro P&Micro R&Micro F$_1$\\ \hline
\emph{Title+Abstract}&11.3&11.3&10.7&47.9&14.2&21.9 \\
\emph{Title+Abstract + ExpSetup}&20.8&20.1&19.4&50.0&23.7&32.2 \\
\emph{Title+Abstract + TableInfo}&29.6&29.1&28.1&68.6&40.3&50.8 \\
\emph{Title+Abstract + ExpSetup + TableInfo}&\textbf{62.5}&\textbf{75.2}&\textbf{65.3}&\textbf{60.8}&\textbf{76.8}&\textbf{67.8} \\
 \hline
\end{tabular}
\end{small}
\end{center}
\caption{\label{tab:exp1ablation}  Ablation experiments results of  \emph{TDMS-IE} for \emph{Task + Dataset + Metric} prediction.}
\end{table*}

\begin{table*}
\begin{center}
%\begin{small}
\small
\begin{tabular}{l cccc cc }
\hline
Task:Dataset:Metric& P@1&P@3&P@5&P@10&\#Correct Score&\#Wrong Task\\ \hline
\emph{Dependency parsing:Penn Treebank:UAS}&1.0&1.0&0.8&0.9&2&0 \\
\emph{Summarization:DUC 2004 Task 1:ROUGE-2}&0.0&0.67&0.8&0.7&0&0 \\
\emph{Word sense disambiguation:Senseval 2:F1}&0.0&0.0&0.1&0.1&0&0 \\
\emph{Word sense disambiguation:SemEval 2007:F1}&1.0&1.0&0.8&0.7&1&0 \\
\emph{Word segmentation:Chinese Treebank 6:F1}&1.0&0.67&0.4&0.2&0&2 \\
\emph{Word Segmentation:MSRA:F1}&1.0&0.67&0.6&0.7&2&3 \\
\emph{Sentiment analysis:SST-2:Accuracy}&1.0&0.67&0.6&0.3&0&3 \\
\emph{AMR parsing:LDC2014T12:F1 on All}&0.0&0.67&0.4&0.2&0&5 \\
\emph{CCG supertagging:CCGBank:Accuracy}&1.0&1.0&1.0&0.8&0&1 \\
\emph{Machine translation:WMT 2014 EN-FR:BLEU}&1.0&0.33&0.2&0.1&0&0 \\ \hline
\emph{Macro-average}&0.70&0.67&0.57&0.46&-&- \\

 \hline
\end{tabular}
%\end{small}
\end{center}
\caption{\label{tab:exp2}  Results of \emph{TDMS-IE}  for ten leaderboards  on  \emph{ARC-PDN}.}
\end{table*}

\subsection{Extraction Results on NLP-TDMS}\label{sec:nlptdes_results}
We evaluate our \emph{TDMS-IE} on the test dataset of \emph{NLP-TDMS}.
Table \ref{tab:exp1} shows the results of our model compared to baselines in different 
evaluation settings: \emph{TDM} extraction (Table \ref{tab:exp1}a), \emph{TDM} extraction excluding papers with ``Unknown'' annotation (Table \ref{tab:exp1}b), and \emph{TDMS} extraction excluding papers with ``Unknown'' annotation (Table \ref{tab:exp1}c).

\emph{TDMS-IE} outperforms baselines by a large margin in all evaluation metrics for the first two evaluation scenarios, where the task is to extract triples $<$\emph{Task, Dataset, Metric}$>$. On testing papers with at least one \emph{TDM} triple annotation, it achieves a macro F$_1$ score of 56.6 and a micro F$_1$ score of 66.0 for predicting \emph{TDM} triples, versus the 37.3 macro F$_1$, and 33.6 micro F$_1$ of the multi-label classification approach.

However, when we add the \textit{score} extraction (TDMS), even if \emph{TDMS-IE} outperforms the baselines, the overall performances are still unsatisfactory, underlining the challenging nature of the task. A qualitative analysis showed that many of the errors were triggered by the noise from the table parser, e.g., failing to identify bolded numeric scores or column headers (see Table \ref{tab:tableExt}). Sometimes a few papers bold the numeric scores for methods from the previous work when comparing to the state-of-the-art results, and our model wrongly predicts these bolded scores for the targeting \emph{TDM} triples.

%
%\subsection{IE Results on NLP-TDMS}\label{sec:nlptdes_results}
%We evaluate our \emph{TDMS-IE} on the test dataset of \emph{NLP-TDMS}.
%Table \ref{tab:exp1} shows the results of our model compared to baselines in different 
%evaluation settings: \emph{TDM} extraction (Table \ref{tab:exp1}a), \emph{TDM} extraction excluding papers with ``Unknown'' annotation (Table \ref{tab:exp1}b), and \emph{TDMS} extraction excluding papers with ``Unknown'' annotation (Table \ref{tab:exp1}c).
%
%\emph{TDMS-IE} outperforms baselines by a large margin in all evaluation metrics for the first two evaluation scenarios. On testing papers with at least one \emph{TDM} triple annotation, it achieves an Macro F$_1$ score of 56.6 and an Micro F$_1$ score of 66.0 for predicting \emph{TDM} triples.
%
%However, the performance of \emph{TDMS-IE} on \emph{TDMS} extraction is still unsatisfactory. This is partially due to the noise from the table parser, e.g., failing to identify bolded numeric scores or column headers. Sometimes a few papers bold the numeric scores for methods from the previous work when comparing to the state-of-the-art results, and our model wrongly predicts these bolded scores for the targeting \emph{TDM} triples.
%

\subsection{Ablations}\label{sec:nlptdes_ablation}
To understand the effect of \emph{ExpSetup} and \emph{TableInfo} in 
document representation \emph{DocTAET} for predicting \emph{TDM} triples, we carry out an ablation experiment. We train and test our system with \emph{DocTAET} containing only \emph{Title+Abstract}, \emph{Title+Abstract+ExpSetup}, and \emph{Title+Abstract+TableInfo} respectively.
Table \ref{tab:exp1ablation} reports the results of different configurations for \emph{DocTAET}.
We observe that both \emph{ExpSetup} and \emph{TableInfo} are helpful for predicting 
\emph{TDM} triples. %In addition, it seems 
It also seems that descriptions from table captions and headers (\emph{TableInfo}) are more informative than descriptions of experiments (\emph{ExpSetup}).

\subsection{Results on ARC-PDN}\label{sec:aclanth_results}
To test whether our system can support %the construction of
to construct various leaderboards from a large number of NLP papers, we apply our model trained on the \emph{NLP-TDMS} training set to \emph{ARC-PDN}. We exclude five papers which also appear in the training set and predict \emph{TDMS} tuples for each paper. 

The set of 77 candidate \emph{TDM} triples comes from the training data, and many of these contain datasets that appear only after 2015.
Consequently, fewer papers are tagged with these triples. 
Therefore, for evaluation we manually choose ten \emph{TDM} triples among all \emph{TDM} triples with at least ten associated papers. These ten \emph{TDM} triples cover various research areas in NLP and contain datasets appearing before 2015. For each chosen \emph{TDM} triple, we rank predicted papers according to the confidence score from the \emph{DocTAET-TDM} model and manually evaluate the top ten results. 

Table \ref{tab:exp2} reports P@1, P@3, P@5, and P@10 for each leaderboard (i.e., \emph{TDM} triple). The  macro average P@1 and P@3 are 0.70 and 0.67, respectively, which is encouraging. %quite good.
 Overall, 86\% of papers 
are related to the target task \emph{T}. We found that most false positives are due to the fact that these papers conduct research on the target task \emph{T}, but report results on a different dataset 
%Some others 
or use the target dataset \emph{D} as a resource to extract features. For instance, most predicted papers for the leaderboard $<$\emph{Machine translation, WMT 2014 EN-FR, BLEU}$>$ are papers about \emph{Machine translation} but 
these papers report results on the dataset \emph{WMT 2012 EN-FR} or \emph{WMT 2014 EN-DE}. 

For \emph{TDMS} extraction, only five extracted \emph{TDMS} tuples are correct. This is a challenging task and more efforts are required to address it in the future.

%even for humans who are not the domain experts. 

% Among the ten leaderboards, results on \{\emph{Word sense disambiguation:Senseval 2:F1} \} and \{\emph{Machine translation:WMT 2014 EN-FR:BLEU}\} are the worst. 

%\subsection{Results in the Zero-shot Setup}\label{sec:zeroshot_results}

\section{Zero-shot \emph{TDM} Classification}\label{sec:zeroshot_results}
Since our framework in principle captures the similarities between \emph{DocTAET} and various \emph{TDM} triples, we estimate that it can perform zero-shot classification of new \emph{TDM} triples at test time. 

We split \emph{NLP-TDMS (Full)} into the training/test sets. The training set contains 210 papers with 96 (distinctive) \emph{TDM} triple annotations and the test set contains 108 papers whose \emph{TDM} triple annotations do not appear in the training set. We train our \emph{DocTAET-TDM} model on the training set as described in Section \ref{sec:tdms-ie-train}. At test time, we use all valid \emph{TDM} triples from \emph{NLP-TDMS (Full)} to form the hypothesis space. %To save the inference time,
To improve efficiency, one could also reduce this hypothesis space by focusing on the related \emph{Task} or \emph{Dataset} mentioned in the paper. 

On the test set of zero-shot \emph{TDM} pairs classification, our model achieves a macro F$_1$ score of 41.6 and a micro F$_1$ score of 54.9, versus the 56.6 macro F$_1$, and 66.0 micro F$_1$ of the few-shot \emph{TDM} pairs classification described in Section \ref{sec:nlptdes_results}. 

%%% Local Variables: 
%%% mode: latex
%%% TeX-master: "nlp-leaderboard"
%%% End: 

\section{Conclusions}\label{sec:con}

In this paper, we have reported a framework %called \emph{TDMS-IE} 
to automatically extract tasks, datasets, evaluation metrics and scores from a set of published scientific  papers in PDF format, in order to reconstruct the leaderboards for various tasks.
We have proposed a method, inspired by natural language inference, to facilitate 
%a joint learning of association 
learning similarity patterns 
between labels and the content words of papers. Our first model extracts $<$\emph{Task, Dataset, Metric}$>$ (\emph{TDM}) triples, and our second model associates the best score reported in the paper to the corresponding \emph{TDM} triple. 
We created two datasets in the NLP domain to test our system. 
Experiments show that our model outperforms the baselines by a large margin in  the identification of \emph{TDM} triples. 

%In the future, more effort is needed in the extraction of the best score. 
In the future, more effort is needed to extract the best score. 
Also the work reported in this paper is based on a small \emph{TDM} taxonomy, we plan to construct a \emph{TDM} knowledge base and provide an applicable system for a wide range of NLP papers.

%To the best of our knowledge, our work is the first attempt towards the creation of NLP Leaderboards in an automatic fashion. The processed datasets and code are publicly available at: \url{https://github.com/IBM/science-result-extractor}.

\section*{Acknowledgments}
The authors appreciate the valuable feedback from the anonymous reviewers.

\bibliographystyle{acl_natbib}
\bibliography{acl2019}

\end{document}